\documentclass[runningheads]{llncs}
\usepackage{amssymb}
\usepackage[T1]{fontenc}
\usepackage{graphicx}
\usepackage{multirow}
\usepackage{makecell}

\begin{document}

\title{ReLA: Representation Learning and Aggregation for Job Scheduling with Reinforcement Learning}

\titlerunning{ReLA: Representation Learning and Aggregation}

\author{Zhengyi Kwan\inst{1,2} \and
Wei Zhang\inst{1}\thanks{Corresponding author.} \and
Aik Beng Ng\inst{2} \and \\Zhengkui Wang\inst{1} \and Simon See\inst{2}}

\authorrunning{Z. Kwan et al.}

\institute{Singapore Institute of Technology, Singapore 828608\\
\email{\{wei.zhang, zhengkui.wang\}@singaporetech.edu.sg}
\and
NVIDIA AI Technology Center, Singapore 038988\\
\email{\{kwanz, aikbengn, ssee\}@nvidia.com}
}

\maketitle            

\begin{abstract}
Job scheduling is widely used in real-world manufacturing systems to assign ordered job operations to machines under various constraints. Existing solutions remain limited by long running time or insufficient schedule quality, especially when problem scale increases. In this paper, we propose ReLA, a reinforcement-learning (RL) scheduler built on structured \underline{re}presentation \underline{l}earning and \underline{a}ggregation. ReLA first learns diverse representations from scheduling entities, including job operations and machines, using two intra-entity learning modules with self-attention and convolution and one inter-entity learning module with cross-attention. These modules are applied in a multi-scale architecture, and their outputs are aggregated to support RL decision-making. Across experiments on small, medium, and large job instances, ReLA achieves the best makespan in most tested settings over the latest solutions. On non-large instances, ReLA reduces the optimality gap of the SOTA baseline by 13.0\%, while on large-scale instances it reduces the gap by 78.6\%, with the average optimality gaps lowered to 7.3\% and 2.1\%, respectively. These results confirm that ReLA’s learned representations and aggregation provide strong decision support for RL scheduling, and enable fast job completion and decision-making for real-world applications.

\end{abstract}

\section{Introduction}
\label{sec:intro}
Manufacturing is a critical economic pillar, which accounts for nearly one fifth of Singapore's gross domestic product (GDP) and over a quarter of China's GDP \cite{edb_sg_manu}. Flexible job shop scheduling (FJSP) is a core optimization problem in modern manufacturing systems, where efficient task assignment and machine utilization directly affect productivity and cost and ineffective scheduling can lead to significant losses \cite{zhang2019review}. FJSP arises in practical scenarios where jobs consist of sequences of operations that must be assigned to compatible machines under precedence and resource constraints. The objective is typically to minimize the overall completion time, \emph{makespan}. The problem's combinatorial nature makes it challenging to solve optimally at scale \cite{meng2020mixed}. As manufacturing systems become more complex and demand real-time decisions, there is a growing need for effective and fast scheduling methods. 

FJSP scheduling has been an active research topic and we group existing studies into two broad stages. In the first stage, solutions center on exact solvers and human-crafted heuristics. A common strong reference in this stage is OR-Tools \cite{ortools}. It uses combinatorial search to generate (near-)optimal schedules, but is not practical for real-world settings due to high high runtime cost. Heuristics such as tabu search, simulated annealing, and genetic algorithms are effective at embedding human expertise and enforcing feasibility, yet they remain impractical because of the requirements of manual design, intensive tuning, and limited adaptability as problem size grows. The limitations of the first stage solutions motivate the research in the second stage, which is dominated by machine learning, especially reinforcement learning (RL). RL is native to FJSP as scheduling is a sequential decision process where the optimal choice depends on the evolving schedule state. Selecting a feasible operation-machine pair directly maps to the RL action space and minimizing makespan aligns with reward-driven policy optimization. Several recent works build on the RL formulation for FJSP, but emphasize different RL aspects. In HGNN \cite{song2022flexible}, the RL state is constructed using graph embeddings that model relations among job operations and machines. LUCA \cite{yang2025llm} augments the state representation by adding an LLM-generated contextual embedding for a variant of FJSP to minimize both makespan and carbon mission. DANIEL \cite{wang2023flexible} represents the state-of-the-art (SOTA) solution. It focuses on learning an attention-based state representation and using the latest proximal policy optimization with the clipped surrogate objective (PPO-Clip) \cite{huang2024ppo} training for decision optimization. While RL models enable fast FJSP scheduling, they remain limited as states are often overly simplified and the scheduling complexity and dynamics are not well captured. This motivates comprehensive representations and effective architectural design for RL scheduling. 

In this paper, we propose ReLA, \underline{re}presentation \underline{l}earning and \underline{a}ggregation for FJSP scheduling with RL. ReLA adopts a representation-first learning philosophy. Instead of relying on a single scheduling representation, ReLA learns multiple representations from scheduling entities, including job operations and machines. ReLA first extracts intra-entity representations by modeling interactions among entities of the same type using two learning modules, including self-attention to capture long-range dependencies, and convolution to capture local structural patterns. It then constructs inter-entity representations using a cross-attention module applied over feasible operation-machine assignment pairs to learn the quality of each pair. To enrich scheduling with a statistical view, ReLA incorporates a mean-pooled representation computed over current entity representations. ReLA aims to better preserve information produced at different abstraction levels, and adopts a multi-scale learning architecture \cite{kwan2025nutrition}, where representations from multiple scales are utilized together as representation guidance. Scheduling decisions are produced by parallel actor networks, which score feasible operation-machine assignments using raw scheduling data, learned representations, and pooled context of multiple scales. In contrast, critic networks use pooled context only to evaluate the overall scheduling state. 

We conduct extensive experiments to evaluate ReLA’s performance. We follow a common practice in scheduling studies and treat OR-Tools as a (near-)optimal benchmark reference, which is effective for quality comparison but too slow for practical use. Across synthetic datasets with small and medium instance sizes with 10–40 jobs, ReLA achieves an average optimality gap of 7.3\% relative to OR-Tools, and reduces the gap of the current SOTA method DANIEL by 13.0\%. On large-scale synthetic instances with 100 jobs or more, ReLA obtains an average gap of 2.1\% with a gap reduction of 78.6\% compared with the SOTA. Such large scales are rarely evaluated in existing studies, yet they reflect realistic industry workloads in domains like smart manufacturing where hundreds of jobs and machines must be scheduled. On public datasets, ReLA achieves a 27.8\% lower average gap than DANIEL. We also show that ReLA performs the best when aggregating diverse learned representations. Overall, ReLA is learning-based and supports fast algorithm execution and shorter job completion times, and is suitable for real-world job scheduling in various applications.

The remainder of this paper is organized as follows. We first introduce the FJSP formulation and scheduling preliminaries in Section \ref{sec:problem}. Then, we present ReLA's technical details in Section \ref{sec:method}. Section \ref{sec:experiment} provides experimental results and discussions. Finally, Section \ref{sec:conclusion} concludes the paper and outlines future directions.

\section{RL Framework for FJSP}
\label{sec:problem}
FJSP is a classical combinatorial optimization problem that can be naturally formulated as a sequential decision-making process. In this study, we adopt an RL framework for FJSP to define the scheduling problem and learning objective. A brief description of the FJSP and the RL formulation is as follows.

\subsection{FJSP}
The FJSP considers a set of $n$ jobs $\mathcal{J}=\{\mathcal{J}_1,\ldots,\mathcal{J}_n\}$ and a set of $m$ machines $\mathcal{M}=\{\mathcal{M}_1,\ldots,\mathcal{M}_m\}$. Each job $\mathcal{J}_i$ consists of a sequence of $n_i$ operations $\{\mathcal{O}_{i,1},\ldots,\mathcal{O}_{i,n_i}\}$, where the processing order of these operations is constrained by precedence relations. Each operation $\mathcal{O}_{i,k}$ can be processed by one machine, selected from a subset of feasible machines $\mathcal{M}^{i,k}\subseteq\mathcal{M}$ for the operation, and the processing time of $\mathcal{O}_{i,k}$ at $\mathcal{M}_j\in \mathcal{M}^{i,k}$ is $p^j_{i,k}>0$. Machines can process at most one operation at a time, and operations are non-preemptive. A \emph{schedule} specifies the operation-machine assignment and start time for every operation while satisfying machine availability and precedence constraints. The objective of FJSP is to minimize the \emph{makespan} $C_{\max}=\max_{i,k} \{C_{i,k}\}$, where $C_{i,k}$ denotes the completion time of operation $O_{i,k}$. Given a start time $t\geq 0$ at which operation $\mathcal{O}_{i,k}$ begins processing on machine $\mathcal{M}_j$, its completion time is $C_{i,k}=t+p^j_{i,k}$.

\subsection{RL Formulation}
The FJSP scheduling process can be modeled as a Markov decision process (MDP), where decisions are made iteratively to assign operations to machines until all operations are scheduled. This formulation aligns naturally with RL. At each decision step, the RL environment reflects the current scheduling \textit{state}, including the processing status of operations, machine availability, etc. The RL agent selects a feasible scheduling decision as an \textit{action} that induces a state transition. The learning objective of RL is to maximize the expected cumulative reward, which is formulated to align with minimizing the makespan. While this RL formulation is standard and shared by many existing solutions, the effectiveness of the learned \textit{policy} depends on how scheduling information is represented and utilized. In the next section, we focus on representation learning and aggregation, which constitute the methodological contribution of this paper.

\section{ReLA: Representation Learning and Aggregation}
\label{sec:method}
In this section, we present ReLA, our representation learning and aggregation approach for FJSP under the RL framework. ReLA learns informative latent representations from different entities of the scheduling problem and aggregates these representations effectively to support policy learning.

\subsection{Design Overview of ReLA}
\begin{figure}[t]
\centering
\includegraphics[width=0.98\textwidth]{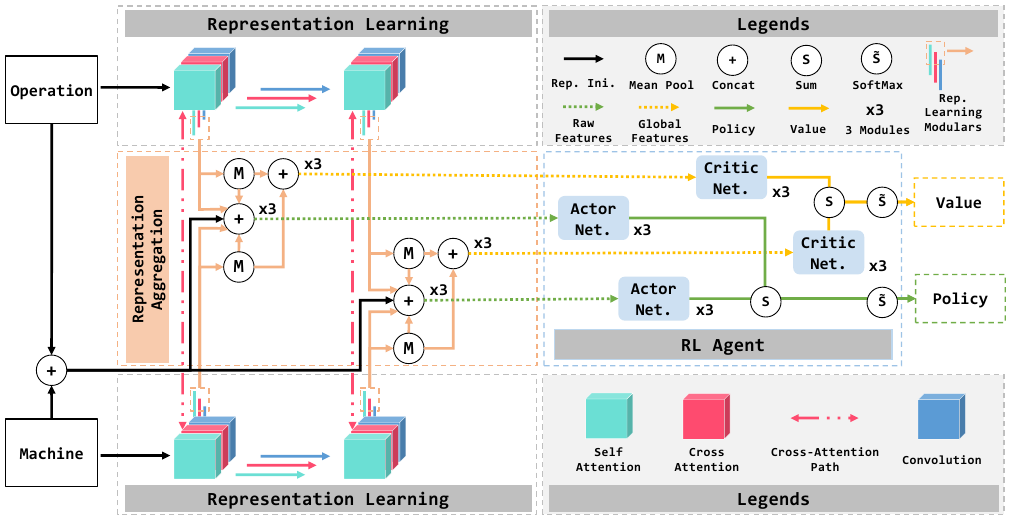}
\caption{ReLA architecture for FJSP scheduling. Multiple representation learning modules extract intra- and inter-entity scheduling representations, which are aggregated to enrich RL decision-making beyond single-representation designs.}
\label{fig:sys}
\end{figure}

Under the RL framework for FJSP, the quality of a schedule depends not only on the framework but also on the representation and utilization of job scheduling information. Existing algorithms, e.g., \cite{wang2023flexible}, often encode the scheduling state using a single representation mechanism, which may not capture the diverse FJSP characteristics sufficiently. In this work, we adopt a representation-centric design for scheduling. We observe that different aspects of FJSP, e.g., operation precedence, machine availability and capability, and operation-machine interactions, exhibit distinct structural properties and therefore benefit from different representation mechanisms. Based on this observation, ReLA follows a two-stage approach. First, multiple latent representations are learned from different entities of the scheduling problem, where each representation emphasizes a particular structural aspect. In the second stage, these learned representations are aggregated to form a decision context that is utilized by the RL policy. Such a learning-aggregation design allows ReLA to exploit diverse and informative representations while avoiding excessive model depth within a single representation space. We illustrate the overall architecture of ReLA in Fig. \ref{fig:sys}, and we then describe the representation learning modules and aggregation strategy in detail.

\subsection{Representation Learning in ReLA}
The first stage of ReLA focuses on learning informative latent representations from different scheduling entities. Instead of constructing a single encoding, ReLA learns multiple representations that capture complementary characteristics of FJSP. As such, ReLA employs dedicated representation learning modules, each of which takes the corresponding raw scheduling information as input and transforms it into a latent representation that emphasizes a specific aspect of the scheduling state. These representations are learned jointly and updated end-to-end through RL. Overall, our expectation is to enable rich and expressive state characterization compared to basic representation approaches. In the following, we describe the detailed representation learning.

\subsubsection{Representation Initialization}
ReLA begins by initializing representations for scheduling entities based on \cite{wang2023flexible}. These initial representations serve as inputs to subsequent representation refinement and interaction modeling stages. For each $\mathcal{O}_{i,k}\in \mathcal{O}$ from job $\mathcal{J}_i$, we construct a feature vector $\mathbf{x}^{\texttt{op}}_{i,k}$ from raw operation information, which describes its processing requirements, execution status, and precedence constraints. The initial representation of $\mathcal{O}_{i,k}$ is obtained as,
\begin{equation}
\mathbf{h}^{\texttt{op}}_{i,k} = \phi_{\texttt{op}}(\mathbf{x}^{\texttt{op}}_{i,k}),
\end{equation}
where $\phi_{\texttt{op}}(\cdot)$ is an embedding function shared across all operations. Similarly, for each machine $\mathcal{M}_j\in \mathcal{M}$, we construct a feature vector $\mathbf{x}^{\texttt{ma}}_j$ from raw machine information, e.g., machine availability, workload, and processing capability. The machine representation is obtained in a similar manner as, $\mathbf{h}^{\texttt{ma}}_j = \phi_{\texttt{ma}}(\mathbf{x}^{\texttt{ma}}_j)$, where $\phi_{\texttt{ma}}(\cdot)$ denotes the machine embedding function. The resulting operation and machine representations $\{\mathbf{h}^{\texttt{op}}_{i,k}\}$ and $\{\mathbf{h}^{\texttt{ma}}_j\}$ serve as the inputs to subsequent ReLA modules.

\subsubsection{Intra-Entity Representation Learning}
With the initial representations, ReLA refines operation and machine representations through intra-entity representation learning. We aim to capture dependencies within each entity type by modeling both global and local patterns in the representation space, and we consider global representation first. ReLA employs a self-attention mechanism \cite{vaswani2017attention} to enable each entity to incorporate information from other entities of the same type. For operations, given the set of operation representations $\{\mathbf{h}^{\texttt{op}}_{i,k}\}$, self-attention allows each operation to aggregate information from other operations through explicit pairwise interactions. Therefore, the mechanism helps capture contextual information related to job operation progression and precedence constraints. The refined operation representations are obtained as,
\begin{equation}
\tilde{\mathbf{h}}^{\texttt{op}}_{i,k} = \texttt{attn}\big(\mathbf{h}^{\texttt{op}}_{i,k}, \{\mathbf{h}^{\texttt{op}}_{i',k'}\}\big),
\end{equation}
where $\texttt{attn}(\cdot)$ denotes a self-attention operator. In this study, the self-attention operator is implemented using a graph attention
layer (GAL) \cite{velivckovic2017graph}, where nodes correspond to entities of the same type and edges encode intra-entity relations. In addition, multi-head attention is adopted to allow different attention heads to capture complementary interaction patterns among entities. Multiple attention heads operate in parallel, and their outputs are aggregated to form the final representation. Similarly, we apply the attention mechanism to machine representations $\{\mathbf{h}^{\texttt{ma}}_j\}$ to extract additional information among machines, such as machine availability and workload distribution. The refined machine representations are obtained as $\tilde{\mathbf{h}}^{\texttt{ma}}_{j} = \texttt{attn}\big(\mathbf{h}^{\texttt{ma}}_{j}, \{\mathbf{h}^{\texttt{ma}}_{j'}\}\big)$. Besides attention-based global patterns, we further consider local patterns to enhance representation expressiveness and stability. In ReLA, we apply a convolutional operator $\texttt{conv}(\cdot)$, which processes information from neighboring representations within a fixed receptive field and produces the local representations as,
\begin{equation}
\hat{\mathbf{h}}^{\texttt{op}}_{i,k} = \texttt{conv}\big(\tilde{\mathbf{h}}^{\texttt{op}}_{i,k},\{\tilde{\mathbf{h}}^{\texttt{op}}_{i',k'}\}\big),\quad \hat{\mathbf{h}}^{\texttt{ma}}_j = \texttt{conv}\big(\tilde{\mathbf{h}}^{\texttt{ma}}_{j},\{\tilde{\mathbf{h}}^{\texttt{ma}}_{j'}\}\big),
\end{equation}
where $\tilde{\mathbf{h}}$ means the global representations for either operations or machines. By sequentially applying self-attention and convolution, intra-entity representation learning in ReLA extracts information through explicit pairwise interactions and localized filtering operations. The produced representations for operations and machines are further processed in the subsequent inter-entity learning.

\subsubsection{Inter-Entity Representation Learning}
After intra-entity representation refinement, ReLA models interactions between different entity types through inter-entity representation learning. We focus on capturing the relationships between operations and machines in this stage, and the relationships are essential for evaluating feasible operation-machine assignments for FJSP scheduling. Specifically, we employ a cross-attention mechanism \cite{jiang2022cross} to enable the interactions between operation representations and machine representations. Given refined representations $\{\hat{\mathbf{h}}^{\texttt{op}}_{i,k}\}$ and $\{\hat{\mathbf{h}}^{\texttt{ma}}_j\}$, cross-attention allows each operation to integrate the machine context into its representation as,
\begin{equation}
\mathbf{z}^{\texttt{op}}_{i,k} = \texttt{c-attn}\big(\hat{\mathbf{h}}^{\texttt{op}}_{i,k}, \{\hat{\mathbf{h}}^{\texttt{ma}}_{j}\}\big),
\end{equation}
where $\texttt{c-attn}(\cdot)$ denotes a cross-attention operator. We modify \texttt{attn} as \texttt{c-attn} where the features of corresponding learning modules are swapped to be used in their counterpart modules. Similarly, we expand machine representations with operation information such as pending operations and execution progress. The interaction-aware machine representations are $\mathbf{z}^{\texttt{ma}}_j = \texttt{c-attn}\big(\hat{\mathbf{h}}^{\texttt{ma}}_j, \{\hat{\mathbf{h}}^{\texttt{op}}_{i,k}\}\big)$. The derived intra-entity representations encode compatibility and interaction information between operations and machines and provide a foundation for subsequent representation aggregation and eventually scheduling decision-making.

\subsubsection{Representation Guidance via Deep Supervision}
To facilitate effective learning of multiple representations, ReLA adopts a two-scale representation learning architecture. Representations produced at both scales are utilized during training as deep supervision \cite{zhang2022contrastive}, rather than relying solely on representations from a single scale. This design allows scheduling information at different levels of abstraction to be preserved and exploited, and enables effective representation learning in the presence of multiple representation modules and scales.

\subsection{Representation Aggregation for Decision-Making}
After representation learning, ReLA aggregates the learned representations to support decision-making. We describe the details of aggregation and RL below.

\subsubsection{Representation Aggregation}
At each decision step $t$, ReLA performs representation aggregation over the current scheduling context. Let $\mathcal{O}_t$ denote the set of unscheduled operations whose precedence constraints are satisfied, and let $\mathcal{M}_t$ denote the set of available machines at step $t$. Representation aggregation is performed over these sets to construct pair-wise representations for candidate operation-machine combinations. ReLA does not collapse all information into a single representation. Instead, for each candidate operation-machine pair $(\mathcal{O}_{i,k}, \mathcal{M}_j)$ for $\mathcal{O}_{i,k}\in\mathcal{O}_t$ and $\mathcal{M}_j\in\mathcal{M}_t$, ReLA constructs aggregated representations based on three representation learning modules, namely self-attention, convolution, and cross-attention. Consider the self-attention as an example. Mean-pooled contextual representations are computed over the current scheduling context for operation $\mathcal{O}_{i,k}$ and machine $\mathcal{M}_j$ as,
\begin{equation}
\tilde{\mathbf{h}}^{\texttt{op,m}} = \texttt{mpool}\big(\{\tilde{\mathbf{h}}^{\texttt{op}}_{i,k} \mid \mathcal{O}_{i,k}\in\mathcal{O}_t\}\big),~\tilde{\mathbf{h}}^{\texttt{ma,m}} = \texttt{mpool}\big(\{\tilde{\mathbf{h}}^{\texttt{ma}}_{j} \mid \mathcal{M}_j\in\mathcal{M}_t\}\big),
\end{equation}
where $\texttt{mpool}(\cdot)$ denotes the mean pooling operator. The operator provides a compact summary of the current scheduling context by aggregating representations across the set of operations or machines. This allows each pair-wise representation to incorporate information beyond the local pair. Together with the raw representations $\mathbf{h}^{\texttt{op}}_{i,k}$ and $\mathbf{h}^{\texttt{ma}}_{j}$, these representations are concatenated to form a self-attention-based aggregated representation for the pair $(\mathcal{O}_{i,k}, \mathcal{M}_j)$ as,
\begin{equation}
\mathbf{h}^{\texttt{attn}}_{i,k,j} =
\texttt{concat}(
\mathbf{h}^{\texttt{op}}_{i,k},
\mathbf{h}^{\texttt{ma}}_{j},
\tilde{\mathbf{h}}^{\texttt{op}}_{i,k},
\tilde{\mathbf{h}}^{\texttt{ma}}_{j},
\tilde{\mathbf{h}}^{\texttt{op,m}},
\tilde{\mathbf{h}}^{\texttt{ma,m}}
).
\end{equation}
Similarly, for the convolution and cross-attention learning modules, we construct aggregated representations as,
\begin{equation}
\mathbf{h}^{\texttt{conv}}_{i,j,k} = \texttt{concat}(\mathbf{h}^{\texttt{op}}_{i,k},\mathbf{h}^{\texttt{ma}}_j,\hat{\mathbf{h}}^{\texttt{op}}_{i,k},\hat{\mathbf{h}}^{\texttt{ma}}_j,\hat{\mathbf{h}}^{\texttt{op,m}},\hat{\mathbf{h}}^{\texttt{ma,m}}),
\end{equation}
\begin{equation}
\mathbf{h}^{\texttt{c-attn}}_{i,k,j} = \texttt{concat}(\mathbf{h}^{\texttt{op}}_{i,k},\mathbf{h}^{\texttt{ma}}_j,\mathbf{z}^{\texttt{op}}_{i,k},\mathbf{z}^{\texttt{ma}}_j,\mathbf{z}^{\texttt{op,m}},\mathbf{z}^{\texttt{ma,m}}),
\end{equation}
where $\hat{\mathbf{h}}^{\texttt{op,m}}$, $\hat{\mathbf{h}}^{\texttt{ma,m}}$, $\mathbf{z}^{\texttt{op,m}}$, and $\mathbf{z}^{\texttt{ma,m}}$ are obtained by applying $\texttt{mpool}(\cdot)$ to the operation and machine representations from the \texttt{conv} and \texttt{c-attn} learning modules, respectively. Overall, at this stage, three aggregated representations $\mathbf{h}^{\texttt{attn}}_{i,k,j}$, $\mathbf{h}^{\texttt{conv}}_{i,k,j}$, and $\mathbf{h}^{\texttt{c-attn}}_{i,k,j}$ are constructed for each candidate operation-machine assignment. These pair-wise aggregated representations characterize the current scheduling context and serve as inputs to the RL described below.

\subsubsection{RL for Scheduling}
Based on the aggregated representations, ReLA formulates the scheduling process within an RL framework. At each decision step $t$, the state $s_t$ is characterized by the current sets of aggregated pair-wise representations,
\begin{equation}
\{\mathbf{h}^{\texttt{attn}}_{i,k,j},\ \mathbf{h}^{\texttt{conv}}_{i,k,j},\ \mathbf{h}^{\texttt{c-attn}}_{i,k,j}\mid \mathcal{O}_{i,k}\in\mathcal{O}_t, \mathcal{M}_j\in\mathcal{M}_t\},
\end{equation}
for all candidate pairs $(\mathcal{O}_{i,k},\mathcal{M}_j)$. For each candidate pair indexed by $(i,k,j)$, ReLA employs multiple actor networks operating in parallel to evaluate the pair-wise representations. These actor networks are organized across two representation scales, refer to Fig. \ref{fig:sys}. At each scale, three actor networks are constructed, corresponding to the three learning modules, i.e., \texttt{attn}, \texttt{conv}, and \texttt{c-attn}. As a result, ReLA employs a total of six actor networks. Each network is implemented as a multilayer perceptron (MLP) network, and let $\texttt{MLP}_{b,s}(\cdot)$ denote the network of module $b\in\mathcal{B}$ with $\mathcal{B}=\{\texttt{attn}, \texttt{conv}, \texttt{c-attn}\}$ and scale $s\in\{1,2\}$. Given index $(i,k,j)$ at $t$, each network takes as input one aggregated representation $\mathbf{h}^{b,s}_{i,k,j}$ at the corresponding scale. Then, the network produces a scalar score as,
\begin{equation}
\ell^{b,s}_t(i,k,j) = \texttt{MLP}_{b,s}(\mathbf{h}^{b,s}_{i,k,j}).
\end{equation}
Accordingly, the final score for the pair is obtained by summing the scalar scores of all actor networks as,
\begin{equation}
\ell_t^{i,k,j} = \sum\nolimits_{s=1}^2 \sum\nolimits_{b\in\mathcal{B}} \ell^{b,s}_t(i,k,j).
\end{equation}
With the computed scores for different feasible operation-machine pairs, we define the action $a_t$ at step $t$ as selecting one pair indexed by $(i,k,j)$. ReLA defines a stochastic policy over the candidate pairs as, 
\begin{equation}
\pi_\theta(i,k,j\mid s_t)
=
\texttt{softmax}\big(\texttt{mask}\big(\ell_t^{i,k,j})\big),
\end{equation}
where $\pi_\theta(a_t|s_t)$ is the stochastic policy to produce $(i,k,j)$ for scheduling. $\texttt{mask}(\cdot)$ assigns $-\infty$ to infeasible operation-machine pairs according to machine availability and processing feasibility, to ensure that only feasible pairs are considered.
While the actor evaluates individual pairs, the critic estimates the overall quality of the current scheduling state in ReLA. Specifically, instead of concatenating six components in the aggregated representation of each learning module, we only consider the pooled representations for intra- and inter-entity representations as,
\begin{equation}
\mathbf{h'}_{\texttt{m}}^{\texttt{attn}} =
\texttt{concat}(
\tilde{\mathbf{h}}^{\texttt{op,m}},
\tilde{\mathbf{h}}^{\texttt{ma,m}}
),~ \mathbf{h'}_{\texttt{m}}^{\texttt{conv}} =
\texttt{concat}(
\hat{\mathbf{h}}^{\texttt{op,m}},
\hat{\mathbf{h}}^{\texttt{ma,m}}
),
\end{equation}
\begin{equation}
\mathbf{h'}_{\texttt{m}}^{\texttt{c-attn}} =
\texttt{concat}(
\mathbf{z}^{\texttt{op,m}},\mathbf{z}^{\texttt{ma,m}}
).
\end{equation}
The corresponding critic networks for all module–scale combinations are modeled as MLPs and together outputs a scalar value estimate at step $t$ as,
\begin{equation}
v_t = \sum\nolimits_{s=1}^2 \sum\nolimits_{b\in\mathcal{B}} v_{\phi_{b,s}}(\mathbf{h'}_{\texttt{m}}^{b,s}),
\end{equation}
where $v_{\phi_{b,s}}$ is the critic network for scale $s$ and learning module $b$. The RL objective in this study is to minimize the makespan, and the reward is designed accordingly. At each decision step $t$, an estimated makespan is computed as the maximum estimated completion time over all operations. The reward $r_t$ is defined as the reduction of this estimate induced by the scheduling decision as,
\begin{equation}
r_t = \max_{\mathcal{O}_{i,k}\in \mathcal{O}} C(\mathcal{O}_{i,k}, s_t) - \max_{\mathcal{O}_{i,k}\in \mathcal{O}} C(\mathcal{O}_{i,k}, s_{t+1}),
\end{equation}
which encourages actions that progressively reduce the estimated makespan and $C(\cdot)$ is the cost function for makespan.

\section{Experiments}
\label{sec:experiment}
In this section, we evaluate ReLA's performance on benchmark FJSP instances and compare it with representative scheduling baselines. 

\subsection{Experimental Setup}

\subsubsection{Datasets}
The FJSP datasets for RL training are generated on the fly during training and the testing datasets are pre-generated with 100 random instances. We adopt two synthetic benchmark data generation schemes, denoted as SD1 and SD2, which are widely used in prior works and implemented using publicly available standard generation code. Each FJSP instance of size $n \times m$ consists of $n$ jobs and $m$ machines, where each job contains a sequence of operations. We follow the procedure in \cite{brandimarte1993routing,song2022flexible} to generate SD1 datasets. The number of operations for each job, the number of compatible machine for each operation, and the processing time of each operation are sampled from uniform distributions, e.g., $\mathcal{U}(1,m)$ for compatible machines and $\mathcal{U}(1,20)$ for processing time. SD2 follows the same generation procedure but uses a wider range of processing times with distribution $\mathcal{U}(1,99)$, which leads to increased variability in operation durations and generally results in more challenging scheduling instances. We consider eight instance sizes for SD1 and SD2, including $10\times5$, $20\times5$, $15\times10$, $20\times10$, $30\times10$, $40\times10$, $100\times10$, and $200\times5$. Among them, the first six are commonly considered in existing studies, where the first four and the following two are considered as small and medium sizes, respectively. The last two represent large-size instances to simulate realistic large-scale scheduling scenarios, and such sizes are not frequently considered in the current literature. Also, we test on public benchmark datasets that are widely used, including the \texttt{mk} instances \cite{brandimarte1993routing} and the \texttt{la} instances \cite{hurink1994tabu}, which are grouped into the \texttt{rdata}, \texttt{edata}, and \texttt{vdata} benchmarks.

\subsubsection{ReLA Configurations}
ReLA emphasizes representation learning and aggregation, and our RL training setup follows existing studies, e.g., \cite{wang2023flexible}. We train all RL models using PPO-Clip \cite{huang2024ppo}. To reduce randomness, all experiments are conducted using five fixed random seeds from 0 to 4, and results are reported as the mean and standard deviation (STD) across multiple runs. We adopt generalized advantage estimation (GAE) to stabilize training and set its parameter $\lambda$ to 0.05 for SD1 and 0.2 for SD2. The attention-based representation learning modules employ 4 attention heads with \texttt{ELU} activation. Each \texttt{conv} operator contains 2 \texttt{Conv1d} layers with kernel size 3, stride 1, padding 1, and a \texttt{Tanh} activation function between the layers. All representation learning modules produce embeddings at two parallel scales, with output dimensions 32 and 8, respectively. The actor and critic networks are optimized using the \texttt{Adam} optimizer with a learning rate of $3\times10^{-4}$. Training is performed for 1{,}000 episodes, during which environment instances are resampled every 20 episodes and policy validation is conducted every 10 episodes. During evaluation, we consider two policy execution modes, namely greedy mode and sampling mode. In greedy mode, the operation–machine pair with the highest policy probability is selected at each decision step. In sampling mode, actions are sampled from the learned stochastic policy. Results for both modes are reported. All experiments are run on an NVIDIA A100 SXM4 80GB GPU.

\subsubsection{Comparison Algorithms}
We evaluate the performance of ReLA and comparison algorithms against a reference solver, OR-Tools \cite{ortools}. OR-Tools is widely used in scheduling research and can produce optimal or near-optimal solutions using advanced combinatorial optimization techniques. However, FJSP is NP-hard, and OR-Tools does not guarantee global optimality within a practical time budget, especially for large-size instances. Therefore, solutions produced by OR-Tools are treated as strong references rather than true optima. We refer to \cite{wang2023flexible} and set a time limit of 1,800 seconds for OR-Tools. We further consider HGNN \cite{song2022flexible} and DANIEL \cite{wang2023flexible} as comparison algorithms. HGNN models FJSP using heterogeneous graph neural networks to capture relationships among scheduling entities. DANIEL is a recent SOTA method that employs attention-based representations for scheduling decision-making. For fairness, whenever possible, we report results directly from the original papers under their recommended settings. In addition, we implement our own versions when results are not available, e.g., for large-size instances. 

\subsection{Experimental Results}

\begin{table}[t]
\centering
\def\arraystretch{1.1}
\setlength{\tabcolsep}{2pt}
\caption{Performance comparison on synthetic datasets SD1 and SD2 with small instance sizes under greedy and sampling modes. Results are reported as makespan and the corresponding gap (\%) relative to OR-Tools. Smaller makespan indicates better performance; the best results are underlined. STDs are reported as subscripts for ReLA.}
\resizebox{0.95\textwidth}{!}{
\begin{tabular}{c|cc|cc|cc|cc}
\hline\hline
Small-Size & \multicolumn{2}{c|}{SD1 $10\times5$} & \multicolumn{2}{c|}{SD1 $20\times5$} & \multicolumn{2}{c|}{SD1 $15\times10$} & \multicolumn{2}{c}{SD1 $20\times10$} \\ \hline
OR-Tools &  96.32 & $-$ & 188.15 & $-$   & 143.53 & $-$  & 195.98 & $-$ \\ \hline
\multicolumn{9}{c}{Greedy Mode} \\ \hline
HGNN  & 111.67 & 15.94\% & 211.22 & 12.26\% & 166.92 & 16.30\% & 215.78 & 10.10\% \\ 
DANIEL & 106.71 & 10.79\% & 197.56 &  5.00\% & 161.28 & 12.37\% & 198.50 &  1.29\% \\ 
ReLA (ours) & \underline{106.64}$_{0.4}$ & \underline{10.71\%} & \underline{195.31}$_{0.4}$ & \underline{3.81\%} & \underline{159.97}$_{0.6}$ & \underline{11.45\%} & \underline{197.79}$_{0.4}$ & \underline{0.92\%} \\ \hline

\multicolumn{9}{c}{Sampling Mode} \\ \hline
HGNN & 105.59& 9.62\%& 207.53 & 10.30\% & 160.86 & 12.07\% & 214.81 &  9.61\% \\ 
DANIEL  & 101.67 &  5.55\% & 192.78 &  2.46\% & 153.22 &  6.75\% & 193.91 & -1.06\% \\ 
ReLA (ours) & \underline{101.05}$_{0.5}$ & \underline{4.91\%} & \underline{191.73}$_{0.3}$ & \underline{1.90\%} & \underline{151.61}$_{0.5}$ & \underline{5.63\%} & \underline{193.65}$_{0.4}$ & \underline{-1.19\%} \\
\hline\hline

\multicolumn{9}{c}{} \\\hline
Small-Size & \multicolumn{2}{c|}{SD2 $10\times5$} & \multicolumn{2}{c|}{SD2 $20\times5$} & \multicolumn{2}{c|}{SD2 $15\times10$} & \multicolumn{2}{c}{SD2 $20\times10$} \\ \hline
OR-Tools & 326.24 & $-$ & 602.04 & $-$ & 377.17 & $-$ & 464.16 & $-$ \\ \hline
\multicolumn{9}{c}{Greedy Mode} \\ \hline
HGNN &  553.61 &  69.69\% & 1059.04 &  75.91\% &  807.47 & 114.09\% & 1045.82 & 126.12\% \\ 
DANIEL &  408.40 &  25.18\% &  671.03 &  11.46\% &  591.21 &  56.75\% &  610.16 &  31.45\% \\ 
ReLA (ours) & \underline{405.50}$_{3.0}$ & \underline{24.29\%} & \underline{663.41}$_{3.2}$ & \underline{10.19\%} & \underline{582.34}$_{1.4}$ & \underline{54.40\%} & \underline{603.96}$_{2.2}$ & \underline{30.12\%} \\ \hline

\multicolumn{9}{c}{Sampling Mode} \\ \hline
HGNN &  483.90 &  48.33\% &  962.90 &  59.94\% &  756.07 & 100.46\% &  990.37 & 113.37\% \\ 
DANIEL &  366.74 &  12.41\% &  629.94 &   4.63\% &  521.83 &  38.35\% & \underline{552.64} & \underline{19.06}\% \\ 
ReLA (ours)  & \underline{361.09}$_{1.3}$ & \underline{10.68\%} & \underline{626.95}$_{1.2}$ & \underline{4.14\%} & \underline{517.81}$_{1.1}$ & \underline{37.29\%} &  562.56$_{1.9}$ &  21.20\% \\ \hline\hline
\end{tabular}
}
\label{tab:small}
\end{table}

\subsubsection{Small-Scale Synthetic Instances}
We evaluate ReLA on small FJSP instances first using SD1 and SD2 datasets. Results are reported under both greedy and sampling modes in Table \ref{tab:small}. First, ReLA achieves the best performance, i.e., the lowest makespan, compared to the comparison algorithm HGNN in all test configurations and DANIEL in fifteen out of sixteen configurations. DANIEL, which is the current SOTA, achieves the best performance on only one configuration, i.e., $20\times 10$ on SD2 with sampling mode, and does not achieve the same level of performance as ReLA in the remaining settings. Second, ReLA's performance gain compared to HGNN is more significant on SD2 datasets, where the processing time distribution exhibits high variability. This suggests that ReLA is effective at handling challenging scheduling conditions. In addition, we observe that sampling mode often leads to improved performance compared to greedy mode. This shows that exploiting the stochasticity during training benefits RL. Notably, in several settings like SD1 $20\times 5$ and $20\times 10$, ReLA in sampling mode is close to, or even better, than OR-Tools, as reflected by small or negative gap percentages. It is worth noting that OR-Tools is executed with a time limit (0.5 hours) and does not guarantee global optimality for difficult instances, e.g., SD2 with high processing time variability. Overall, ReLA's competitive performance on small instances shows that ReLA provides effective scheduling decisions, where its unique representation learning and aggregation framework provides comprehensive information from diverse modules.

\begin{table*}[t]
\centering
\def\arraystretch{1.1}
\setlength{\tabcolsep}{2pt}
\caption{Performance comparison on synthetic datasets with medium and large instance sizes and public datasets under greedy and sampling modes. Results are reported as makespan and the gap (\%) relative to OR-Tools. Smaller makespan indicates better performance; the best results are underlined. STDs are reported as subscripts.}
\resizebox{0.925\textwidth}{!}{
\begin{tabular}{c|cc|cc|cc|cc}
\hline\hline
Medium-Size  & \multicolumn{2}{c|}{SD1 $30\times10$}  & \multicolumn{2}{c|}{SD1 $40\times10$} & \multicolumn{2}{c|}{SD2 $30\times10$}  & \multicolumn{2}{c}{SD2 $40\times10$} \\ \hline
OR-Tools  & 274.67 & $-$  & 365.96 & $-$  & 692.26 & $-$  & 998.39 & $-$ \\ \hline
\multicolumn{9}{c}{Greedy Mode} \\ \hline
HGNN & 314.71 & 14.58\% & 417.87 & 14.18\% & 1564.57 & 126.01\% & 2048.96 & 105.23\% \\ \hline
DANIEL & 288.61 & 5.08\% & 379.28 & 3.64\% & 794.62 & 14.79\% & 983.37& -1.50\% \\ \hline
ReLA (ours) & \underline{284.32}$_{2.2}$ & \underline{3.51\%} & \underline{373.87}$_{2.6}$ & \underline{2.16\%} & \underline{772.53}$_{2.3}$ & \underline{11.60\%} & \underline{958.13}$_{4.7}$ & \underline{-4.03\%} \\ \hline
\multicolumn{9}{c}{\footnotesize Sampling Mode} \\ \hline
HGNN & 308.55 & 12.33\% & 410.76 & 12.24\% & 1486.56 & 114.74\% & 1976.25 & 97.94\% \\ \hline
DANIEL & 286.77 & 4.41\% & 379.71 & 3.76\% & 757.48 & 9.42\% & 951.21 & -4.73\% \\ \hline
ReLA (ours) & \underline{281.79}$_{2.8}$ & \underline{2.59\%} & \underline{372.82}$_{3.8}$ & \underline{1.87\%} & \underline{734.66}$_{1.3}$ & \underline{6.12\%} & \underline{925.82}$_{2.1}$ & \underline{-7.27\%} \\ \hline

\multicolumn{9}{c}{\scriptsize ~}\\ \hline\hline
Large-Size & \multicolumn{2}{c|}{SD1 $100\times10$} & \multicolumn{2}{c|}{SD1 $200\times5$} & \multicolumn{2}{c|}{SD2 $100\times10$} & \multicolumn{2}{c}{SD2 $200\times5$} \\ \hline
OR-Tools & 944.2 & $-$ & 1884.7 & $-$ & 2114.5 & $-$ & 5876.3 & $-$ \\ \hline
\multicolumn{9}{c}{Greedy Mode} \\ \hline
DANIEL & 946.0$_{19.5}$ & 0.19\% & 1901.1$_{9.0}$ & 0.87\% & 2233.4$_{21.7}$ & 5.62\% & \underline{5923.8$_{6.4}$} & \underline{0.81\%} \\ \hline
ReLA (ours) & \underline{914.7$_{4.8}$} & \underline{-3.12\%} & \underline{1898.2$_{18.0}$} & \underline{0.72\%} & \underline{2209.9$_{19.6}$} & \underline{4.51\%} & 6014.1$_{62.4}$ & 2.35\% \\ \hline

\multicolumn{9}{c}{Sampling Mode} \\ \hline
DANIEL & 985.3$_{35.6}$ & 4.35\% & 2018.2$_{23.4}$ & 7.08\% & 2253.2$_{51.4}$ & 6.56\% & 7093.4$_{492.1}$ & 20.71\% \\ \hline

ReLA (ours) & \underline{923.4$_{11.3}$} & \underline{-2.20\%} & \underline{1934.8$_{30.6}$} & \underline{2.66\%} & \underline{2200.2$_{9.5}$} & \underline{4.05\%} & \underline{6099.8$_{92.1}$} & \underline{3.80\%} \\ \hline

\multicolumn{9}{c}{}\\\hline\hline
Public Data  & \multicolumn{2}{c|}{mk}  & \multicolumn{2}{c|}{rdata}  & \multicolumn{2}{c|}{edata}  & \multicolumn{2}{c}{vdata} \\ \hline
OR-Tools & 174.20 & $-$  & 935.80 & $-$  & 1028.93 & $-$  & 919.60 & $-$ \\ \hline
\multicolumn{9}{c}{Greedy Mode} \\ \hline
HGNN & 201.40 & 15.61\% & 1030.83 & 10.15\% & 1187.48 & 15.41\% & 955.90 & 3.95\% \\ \hline
DANIEL & 185.70 & 6.60\% & 1031.63 & 10.24\% & 1194.98 & 16.14\% & \underline{944.85} & \underline{2.75\%} \\ \hline
ReLA (ours) & \underline{185.02}$_{1.6}$ & \underline{6.21\%} & \underline{1025.03}$_{4.1}$ & \underline{9.54\%} & \underline{1179.45}$_{5.5}$ & \underline{14.63\%} & 946.87$_{2.5}$ & 2.97\% \\ \hline

\multicolumn{9}{c}{Sampling Mode} \\ \hline
HGNN & 190.60 & 9.41\% & 985.30& 5.29\%& 1116.68 & 8.53\% & 930.80 & 1.22\% \\ \hline
DANIEL & 180.80 & 3.79\% & \underline{978.28} & \underline{4.54\%} & 1122.60 & 9.10\% & 925.40 & 0.63\% \\ \hline
ReLA (ours) & \underline{179.48}$_{0.5}$ & \underline{3.03\%} & 979.14$_{1.8}$ & 4.63\% & \underline{1120.58}$_{3.9}$ & \underline{8.91\%} & \underline{925.25}$_{0.4}$ & \underline{0.61\%} \\ \hline\hline
\end{tabular}
}
\label{tab:medium-large-pub}
\end{table*}

\subsubsection{Medium-Scale Synthetic Instances}
We further evaluate the performance on medium-size instances using SD1 and SD2 and present the results in Table \ref{tab:medium-large-pub}. For these larger instances, training a separate RL model for each size becomes computationally expensive. Thus we follow common practice in \cite{song2022flexible,wang2023flexible} and directly apply the model trained on $10\times 5$ instances to medium size experiments without retraining. Although the models are not specialized to the instance sizes, ReLA maintains competitive performance consistently, with the lowest makespan across all test configurations. This demonstrates ReLA's generalizability and scalability. For example, on SD1 with size $40\times10$, ReLA reduces DANIEL’s makespan from 379.3 and 379.7 under greedy and sampling modes to 373.9 and 372.8, respectively, corresponding to optimality gap reductions of 40.8\% and 50.4\% relative to OR-Tools. On SD2 with higher scheduling difficulty, ReLA maintains consistent improvements over comparison algorithms, e.g., reduces DANIEL’s makespan from 794.6 to 772.5, a reduction of 22.1, under greedy mode for size $30\times 10$. Although the optimality gap reduction is smaller than that observed on SD1, ReLA remains the most competitive and achieves negative gaps on multiple SD2 settings, meaning better performance than OR-Tools. For instance, on size $40\times 10$, ReLA achieves negative gaps of $-4.0\%$ in greedy mode and $-7.3\%$ in sampling mode relative to OR-Tools. Overall, these results confirm that ReLA achieves robust performance and strong generalization when transferred to different instance sizes without retraining.

\subsubsection{Large-Scale Synthetic Instances}
Large-size instances are rarely considered in existing studies. However, large problem sizes are common and often unavoidable in real-world applications such as smart manufacturing. To bridge this gap, we evaluate ReLA on large-scale instances with sizes $100\times10$ and $200\times5$ and report the results in Table \ref{tab:medium-large-pub}. Same as the above experiments, we use the same model trained on $10\times 5$ instances without retraining. We compare ReLA with DANIEL, where the results are based on our own implementation, which matches DANIEL's performance in small-size instances and allows us to report STD as well. Seen from the table, ReLA remains competitive when directly transferred to sizes far beyond $10\times 5$. Although algorithms experience increased difficulty and variance as instance size grows, ReLA achieves lower makespan relative to DANIEL in most settings. Besides, OR-Tools does not consistently deliver better solutions, although with much higher runtime (30 minutes) which makes it unsuitable for large-scale or time-sensitive scheduling. This reflects the difficulty of FJSP at this scale. Learning-based methods are more efficient, e.g., on SD1 $100\times10$, DANIEL takes 0.07 and 1.28 minutes per instance in greedy and sampling modes, while ReLA needs 0.16 and 1.16 minutes, respectively. Although ReLA incurs higher computation cost than DANIEL in greedy mode due to its diverse representation learning, it exhibits lower runtime in sampling mode. This may indicates that the additional representation overhead can be effectively offset when stochastic action selection is employed. Overall, we can see that ReLA scales well to large instances and owns potential for practical deployment in large-scale industrial applications, where retraining is often impractical.

\subsubsection{Public Datasets}
We evaluate ReLA on two widely used public datasets \texttt{mk} and \texttt{la}, where \texttt{la} includes \texttt{rdata}, \texttt{edata}, and \texttt{vdata}. These datasets exhibit irregular routing and machine-compatibility patterns, and we show results in Table \ref{tab:medium-large-pub}. ReLA outperforms comparison algorithms in most tests and approximates OR-Tools performance closely, e.g., for \texttt{vdata} in sampling mode. This shows that ReLA's representation learning and aggregation generalize beyond procedurally generated synthetic SD1 and SD2 datasets. 

\begin{table*}[t]
\centering
\def\arraystretch{1.1}
\setlength{\tabcolsep}{2pt}
\caption{Ablation study on SD1 $10\times 5$ instances. \texttt{a}, \texttt{x}, \texttt{c}, and \texttt{s} denote the \texttt{attn}, \texttt{c-attn}, and \texttt{conv} learning modules, and deep supervision, respectively. Two variants are based on ReLA. $L$ is the number of scales. Each entry is reported as makespan with the gap (\%) relative to OR-Tools. Smaller makespan indicates better performance; the best results are underlined, including gap values. STDs are reported as subscripts.}
\resizebox{0.95\textwidth}{!}{
\begin{tabular}{c|cccc|cc|cc|cc|cc}
\hline\hline
Model & $\texttt{a}$ & $\texttt{x}$ & $\texttt{c}$ & $\texttt{s}$ & \multicolumn{2}{c|}{$L=8$; Greedy} & \multicolumn{2}{c|}{$L=8$; Sampling} & \multicolumn{2}{c|}{$L=2$; Greedy} & \multicolumn{2}{c}{$L=2$; Sampling} \\ \hline

OR-Tools & $-$ & $-$ & $-$ & $-$ & \multicolumn{8}{c}{96.3} \\ \hline

DANIEL & $\checkmark$ & $\times$ & $\times$ & $\times$ & 109.9$_{0.4}$ & 14.1\% & 103.7$_{0.3}$ & 7.7\% & 108.1$_{0.3}$ & 12.3\% & 102.4$_{0.1}$ & 6.3\% \\ \hline

Variant-A & $\checkmark$ & $\checkmark$ & $\times$ & $\times$ & 108.2$_{0.9}$ & 12.4\% & 102.4$_{0.9}$ & 6.3\% & 107.2$_{0.3}$ & 11.3\% & 101.7$_{0.2}$ & 5.6\% \\ \hline

Variant-B
& $\checkmark$ & $\checkmark$ & $\checkmark$ & $\times$ & 108.4$_{0.9}$ & 12.6\% & 102.0$_{0.7}$ & 5.9\% & 107.8$_{0.4}$ & 11.9\% & 101.6$_{0.5}$ & 5.5\% \\ \hline

ReLA (ours)
& $\checkmark$ & $\checkmark$ & $\checkmark$ & $\checkmark$ & \underline{107.6$_{0.7}$} & \underline{11.7\%} & \underline{101.4$_{0.5}$} & \underline{5.3\%} & \underline{106.6$_{0.4}$} & \underline{10.7\%} & \underline{101.1$_{0.5}$} & \underline{5.0\%} \\ \hline\hline
\end{tabular}
}
\label{tab:ablation}
\end{table*}

\subsubsection{Ablation Study}
We perform an ablation study to evaluate the contribution of ReLA’s multi-scale architecture and representation learning modules, with results shown in Table \ref{tab:ablation}. A key variable is the number of representation scales $L$. The default ReLA model uses two scales as shown in  Fig. \ref{fig:sys}. The configuration enables ReLA to extract and utilize intermediate representations from two scales to support the RL agent in quantifying the quality of each operation–machine pair. When $L$ becomes large (e.g., 8), many scales tend to generate similar representations, and it becomes harder to differentiate and identify the best pairs. Consequently, performance does not improve with higher $L$, and ReLA achieves the lowest makespan under its default two-scale setting. We further validate the contribution of individual learning modules. Variant-A improves over the attention-only SOTA method DANIEL by introducing a \texttt{c-attn} module to learn inter-entity representations. Variant-B further improves over Variant-A by adding a \texttt{conv} module to expand intra-entity representations with convolution. Finally, the full ReLA model performs best, and its strong performance is largely attributed to the combined use of multiple learning modules across multiple scales and effective representation aggregation, which increase representation diversity and supports better scheduling decisions.

\section{Conclusion}
\label{sec:conclusion}
In this paper, we proposed ReLA, an RL scheduler that learns and combines multiple representations from job operations and machines to optimize scheduling decisions. Experiments show that ReLA achieves the best makespan in most test settings and outperforms comparison learning-based solutions. For large-scale instances, ReLA achieves an average optimality gap of 2.1\% relative to OR-Tools, reducing the current SOTA's gap by 78.6\%. This study confirms that learning richer representations across multiple scales and different learning modules provides more effective state representation for RL scheduling compared to simplified representation. Overall, ReLA demonstrates competitive scheduling quality with fast decision-making. Its competitive performance supports solution deployment in smart manufacturing systems with industrial-scale workloads.

\subsubsection{Acknowledgements} 
This work was supported in part by A*STAR under its MTC Individual Research Grants (IRG) (Award M23M6c0113), MTC Programmatic (Award M23L9b0052), Singapore’s Economic Development Board - Industrial Postgraduate Programme, in conjunction with NVIDIA AI Technology Center and SIT, and the National Research Foundation, Singapore and DSO National Laboratories under the AI Singapore Programme (AISG Award No: AISG2-GC-2023-006).

\bibliographystyle{splncs04}
\bibliography{references}

\end{document}